\documentclass[twoside]{article}

\usepackage[preprint]{aistats2026}
\usepackage[round]{natbib}

\usepackage{url} 
\usepackage{booktabs}
\usepackage{amsmath}
\usepackage{amssymb}
\usepackage{amsthm}
\usepackage{graphicx}
\usepackage{algorithmic}
\usepackage{algorithm}
\usepackage{xcolor}

\usepackage{enumitem}
\usepackage{calc}

\usepackage{caption}
\usepackage{subcaption}

\theoremstyle{plain}

\newtheorem{proposition}{Proposition}
\newtheorem{lemma}{Lemma}
\newtheorem{corollary}{Corollary}

\newcommand{\grad}{\nabla}

\newcommand{\tar}{\text{target}}
\newcommand{\sr}{d^\pi}
\newcommand{\som}{\rho^\pi}

\begin{document}

%

%

\twocolumn[

\aistatstitle{Bellman Diffusion Models}

\aistatsauthor{Liam Schramm \And Abdeslam Boularias}

\aistatsaddress{Montanuniversität Leoben \And Rutgers University} 
]





\begin{abstract}
  The state occupancy measure (SOM) and successor state measure (SSM) are important theoretical tools in reinforcement learning that represent the distribution of future states.  However, while these tools see extensive use in theory and theoretically-motivated algorithms, they have not seen significant use in practical settings because existing algorithms for learning SOM and SSM are high-variance or unstable in practice. To address this, we explore using diffusion models as a representation for the SSM. We find that enforcing the Bellman flow constraints on a diffusion model leads to a temporal difference update on the predicted noise, similar to the standard Q-learning update on the predicted reward. As a result, our method has the expressive power of a diffusion model, and a low variance that is comparable to that of TD-learning. To demonstrate this method's practicality, we propose a simple offline reinforcement learning algorithm based on regularizing the learned SSM. We test the proposed method on an array of offline RL problems, and find it has the highest average performance of all methods in the literature, as well as achieving state-of-the-art performance on several environments. 
\end{abstract}

\section{Introduction}

The state occupancy measure (SOM) and successor state measure (SSM) are common objects of study in reinforcement learning (RL). A common statement of the objective in RL is to find a policy that induces the state occupancy measure with the highest expected reward \citep{lee2020efficientexplorationstatemarginal, lee2021optidiceofflinepolicyoptimization, ma2022farillgooffline, ma2022versatileofflineimitationobservations, nachum2020reinforcementlearningfenchelrockafellarduality}. The state occupancy measure  has also received considerable attention in the RL theory community, as a number of provably efficient exploration schemes revolve around regularizing the state occupancy measure \citep{dann2023bestworldspolicyoptimization, jin2023improvedbestofbothworldsguaranteesmultiarmed, zimmert2022tsallisinfoptimalalgorithmstochastic}. The successor state measure (SSM) is a closely related concept, which describes the distribution over future states, given that the agent is currently at state $s$ and takes action $a$.

These tools have been of particular interest in offline RL, where a central problem is keeping the future state distribution of a policy within the support of the dataset. For this reason, a large number of works have formulated the learning problem as an attempt to imitate the state distribution of the dataset \citep{lee2021optidiceofflinepolicyoptimization, ma2022versatileofflineimitationobservations, ho2016generativeadversarialimitationlearning}. However, it is difficult to implement this expression of the problem as a learning objective directly, for a number of reasons. If we learn a generative model of the state distribution, it is not clear how to extract the optimal policy from the model. Learning the probability of a given state by regression is challenging, because the learned function does not typically integrate to 1 and so is not a valid probability distribution. For this reason, most other works are forced to use a heavy set of mathematical tricks, based on Fenchel convex dual functions for example, to arrive at a weighted behavior cloning objective. This adds a great deal of technical overhead when adapting the problem to new settings \citep{ma2022versatileofflineimitationobservations, nachum2019algaedicepolicygradientarbitrary}. Additionally, the weighted behavior cloning objectives that result from this approach often struggle to match the performance of the actor-critic methods like TD3-BC ~\citep{park2024valuelearningreallymain}.

We propose {\it Bellman Diffusion Models}, the first SSM estimator that meets the following desideratum.
\begin{enumerate}
	\item The proposed SSM estimator is off-policy, so it can be learned for arbitrary policies offline. 
	\item The proposed SSM estimator is generative, so it can be sampled from. 
	\item An upper bound on the KL divergence between the proposed SSM estimator and arbitrary continuous distributions is easy to calculate.
\end{enumerate}
The combination of these three factors means that it can be calculated and used as part of a regularization term for the primal RL problem, without needing to derive a Fenchel dual formulation, which considerably simplifies its use. We also theoretically analyze this approach, and provide similar convergence guarantees to those that exist for deep Q-learning algorithms.

We present our algorithm for learning the Bellman Diffusion Model (BDM). We then additionally propose an offline RL algorithm called ReBRAC with State Behavior Cloning (TD3-SBC). TD3-SBC starts with ReBRAC (A variant of TD3-BC based on TD3 + a behavior cloning term), then learns a Bellman Diffusion Model and uses it to regularize the divergence between the SSM of the policy and the future trajectory of the state. This effectively encourages the agent to clone the state distribution as well as the action distribution, preventing distribution shift.  We find that this method achieves  state-of-the-art results on several offline RL tasks and also has the highest average performance of any method in the literature. To summarize, our contributions are as follows: 
\begin{enumerate}
	\item We present Bellman Diffusion Models (BDM), a successor state measure estimator that makes it possible to solve the simpler primal formulation of offline RL problems, instead of the more challenging dual formulation
	\item We show that BDMs have the correct distribution as a fixed point of their update rule, which is the same guarantee given for common deep value-learning algorithms
	\item We propose an offline RL algorithms based on regularizing the successor state measure, and show it achieves state of the art results.
\end{enumerate}

In addition to our empirical results, we hope that this work lays the foundation for future work based on explicit regularization of the successor state measure and helps to bridge the gap between more traditional RL algorithms and works focused on state occupancy.

\section{Background}
	Diffusion models are a form of generative model that has shown significant success in image generation \citep{ho2020denoisingdiffusionprobabilisticmodels}. 
	In our work, we are primarily concerned with the loss function of diffusion models, and how it can be used to derive a Bellman update. For this reason, we begin with a review of diffusion models and the derivation of the standard diffusion model loss. Diffusion models are trained using a forward process and a backward process. In the forward process, noise is gradually added to a data point until only noise remains, and the data point is distributed as a multivariate unit Gaussian. If the noise is added successively over $K$ steps, then this produces a sequence of increasingly random points from $x_0$ (a random point from the dataset $D$) to $x_K$. 
    
    We present an overview of the results from Denoising Diffusion Probabilistic Models (DDPM)\citep{ho2020denoisingdiffusionprobabilisticmodels}. 
    Let $D$ be a dataset and $x_0$ be a data point in $D$. The probability of a sequence of noised points $x_{0:K}$ is then
	$$q(x_{0:K}) = q(x_0) \prod_{i=1}^K q(x_{i} | x_{i-1}, x_0),$$
    $q(x_0)$ is defined to be $\frac{1}{| D |}$ for each point in $D$ and 0 for all other $x_0$. For all other time steps, the forward process probabilities are 
    $$q(x_i | x_{i-1}) = \mathcal{N}(\sqrt{1-\beta_i}x_{i-1}, \beta_i I),$$
    where $\beta_i$ is the forward variance at the $i^{\textrm{th}}$ step. An advantage of the use of Gaussian noise is that it allows a closed form solution for the distribution after $i$ steps, because the sum of all of additive Gaussian noises up to step $i$ is also Gaussian. If we define $\alpha_i = 1-\beta_i$ and $\Bar{\alpha_i} = \prod_{j=0}^i \alpha_j$, then 

    $$q(x_i | x_0) = \mathcal{N}(\sqrt{\Bar{\alpha_i}}x_0, (1-\Bar{\alpha_i}) I)$$

	In the reverse process, a neural network parameterized by weights $\theta$ outputs a Gaussian distribution with mean $\epsilon_\theta$, predicting the noise that was added during the forward process. The backward process samples a predicted noise from this distribution and this noise is subtracted from the data point. This process repeats for the same number of steps as the forward process. The probability of a sequence of points $x_{0:K}$ in the reverse diffusion process is 
	$$p(x_{0:K} | \theta) = p(x_K) \prod_{i=1}^{K} p(x_{i-1} | x_{i}, \theta).$$

	The diffusion model is trained to minimize the evidence lower bound. \citet{ho2020denoisingdiffusionprobabilisticmodels} derive the following loss as an upper bound to the negative log probability of the data.\footnote{Unlike \citet{ho2020denoisingdiffusionprobabilisticmodels}, we assume the forward and backward process variances are the same.}
    
    \begin{align*}
        E_{x \sim D} &[- log(p_\theta(x))] \leq \\&(K-1) E_{x_i, i}\left[
        \frac{1-\bar{\alpha}_i}{2 (1-\bar{\alpha}_{i-1}) \beta_i}
        || \Tilde{\mu}(x_i, x_0) - \mu_\theta(x_i, i) ||^2 \right]
    \end{align*}

    where $i \sim Uniform(1, K), \epsilon \sim \mathcal{N}(0, I), \text{ and } x_i = \sqrt{\Bar{\alpha_i}}x_0 + \sqrt{1-\Bar{\alpha_i}}\epsilon$. This can be reparameterized as follows, so that the neural network outputs $\epsilon_\theta$, an estimate of the noise $\epsilon$ added to the original sample.  
    
    \begin{align*}
        E_{x \sim D} &[- log(p_\theta(x))] \leq \\&(K-1)E_{i, x_0, \epsilon} 
        \left[\frac{\beta_i}{2 \alpha_i (1-\Bar{\alpha_i})}|| \epsilon - \epsilon_\theta(x_i, i) ||^2 \right]
    \end{align*}

\section{Related work}
    {\bf Diffusion models.} 
    Diffusion models have seen significant success as a class of generative models for image synthesis \citep{ho2020denoisingdiffusionprobabilisticmodels, dhariwal2021diffusionmodelsbeatgans}. More recently, there has been a growing interest in using diffusion models for imitation learning and reinforcement learning, especially to represent policies. Diffusion planners propose a model in which denoising is analogous to planning, and perform trajectory optimization by denoising \citep{janner2022planningdiffusionflexiblebehavior}. 
        Diffusion Policies extend this approach to behavior cloning \citep{chi2023diffusionpolicy}. Diffusion Q-learning proposes using diffusion models as an expressive class of policies for offline learning \citep{wang2023diffusionpoliciesexpressivepolicy}. 
        
    {\bf Learned successor state measures.} 
    Successor representations learn the distribution of future states, given the current state and action \citep{dayan1993successorrepresentations}. $\gamma$-models generalize this idea to continuous state distributions by learning a generative representation of future states \citep{janner2021generativetemporaldifferencelearning}. Our method differs from $\gamma$-models in that $\gamma$-models only permit a low-variance score matching loss under special circumstances, when the log probability of the future state distribution can be directly calculated under the model, such as normalizing flows, and the environment dynamics are deterministic. By contrast, our method permits this kind of low-variance backup works for stochastic environments and for diffusion models, which are much more expressive than normalizing flows.

	In contemporaneous work, ~\citet{farebrother2025temporaldifferenceflows} present an alternate derivation of Bellman Diffusion Models ($TD^2-DD$ in their terminology). Our work differs in several regards. First, ~\citet{farebrother2025temporaldifferenceflows} focus only on prediction of the SSM -- they do not explore its use in offline RL, or attempt to learn policies by backpropagating through the model. While their theoretical results are more general, a stated goal of this work is develop an approach to state regularization that does not require extensive mathematical knowledge outside of RL (such as convex analysis or stochastic calculus) to understand or implement. Our theoretical results are consistent with this goal, using only Jensen's inequality and standard properties of the KL divergence to prove our results. In contrast, ~\citet{farebrother2025temporaldifferenceflows} use a substantial amount of stochastic calculus, making it difficult to understand for researchers with solely an RL background. 

    {\bf Successor state measure and state occupancy measure in imitation learning and offline reinforcement learning.} 
    GAIL frames the problem of imitation learning as a problem of state-occupancy matching. It then solves this problem by learning a cost function that maximally separates the real data from the policy data. Minimizing this cost causes the policy to imitate the expert \citep{ho2016generativeadversarialimitationlearning}. Unlike some other methods, GAIL is online and assumes that the agent has access to the environment. Another approach to state occupancy matching in imitation learning and offline reinforcement learning is the DICE family of algorithms. 
    AlgaeDICE poses the offline reinforcement learning problem as finding the state occupancy measure with the highest expected reward \citep{nachum2019algaedicepolicygradientarbitrary}. It solves this by first applying a Fenchel transform, and then solving the dual problem. In practice, this results in a value estimation problem, with weighted behavior cloning for the policy.   SMODICE takes a similar approach to offline imitation learning, applying a Fenchel transform, learning a value function, and using the value function to produce a weighted behavior cloning method \citep{ma2022versatileofflineimitationobservations}

    {\bf General Utilities and Convex RL.} 
	A related field of is General Utilities or Convex RL, in which the reward function can be a nonlinear function of the occupancy measure. This is relevant to a number of problems, including offline learning and exploration. \citet{zhang2020variationalpolicygradientmethod} derive a Variational Policy Gradient Theorem for RL with general utilities, analogous to the Policy Gradient theorem for RL with cumulative reward. \citet{mutti2021taskagnosticexplorationpolicygradient} propose a state entropy objective for exploration, and optimizes it using a k-nearest-neighbor that avoids directly calculating the state distribution. \citet{desanti2025provablemaximumentropymanifold} propose a method for maximum entropy exploration with diffusion models.

\section{Notation and Definitions}
\label{sec:notation}
\begin{description}[leftmargin=!,labelwidth=\widthof{\bfseries $L_{\mathrm{BDM}}$}]
  \item[$S$] State space, subset of $\mathbb{R}^{dim(S)}$.
  \item[$A$] Action space, subset of $\mathbb{R}^{dim(A)}$.
  \item[$s$] State variable, element of state space $S$.
  \item[$a$] Action variable, element of action space $A$.
  \item[$\pi(a\!\mid\!s)$] Policy: mapping $s\in\mathcal{S}$ to a distribution over $A$.
  \item[$T(s'\!\mid\!s,a)$] Transition kernel: probability density of next state $s'\in S$ given $(s,a)$.
  \item[$s'$] State variable, element of state space $S$, referring to the state found after taking action $a$ in state $s$.
  \item[$\gamma$] Discount factor, scalar in $[0,1)$.
  \item[$d^{\pi}(s_f \!\mid\! s, a)$] Discounted state‑occupancy measure under policy $\pi$, subject to Bellman Flow Constraints, \ 
    $d^{\pi}(s_f | s, a) = (1-\gamma) T(s' = s_f | s, a) + \gamma E_{a' \sim \pi(s'),  s' \sim T( \cdot | s, a)}[d^{\pi}(s_f | s', a')]$
  \item[$Q_{\pi}(s,a)$] Action‑value function under policy $\pi$, 
    $Q_{\pi}(s,a)=E\bigl[\sum_{t=0}^\infty \gamma^{t}r_{t}\mid s_{0}=s,a_{0}=a\bigr]$.
  \item[{$K$}] {Number of diffusion steps; positive integer.}
  \item[{$i$}] {Diffusion timestep index, integer in $\{1,\dots,K\}$.}
  \item[{$x_{0}$}] {Clean future‑state sample, $x_{0}\in\mathcal{S}$.}
  \item[{$x_{i}$}] {Noised sample at step $i$, $x_{i} = \sqrt{\bar{\alpha}_{i}}x_0 + \sqrt{1-\bar{\alpha}_{i}} \epsilon$.}  
  \item[{$\mu_\theta(x_{0}, i)$}] {Predicted mean of the reverse diffusion Gaussian: the model’s estimate of the mean of $x_{i-1}$ given the current noised sample $x_{i}$, parameterized by $\theta$.}
  \item[{$\mu_\theta(s_{f_i}, i, s, a)$}] {The Bellman diffusion model’s estimate of the reverse-diffusion mean: given the noised future state $s_{f_i}$ at timestep $i$ (conditioned on $(s,a)$),$\mu_\theta$ predicts the Gaussian mean of the previous sample $s_{f_{i-1}}$.}
  \item[{$\epsilon$}] {Gaussian noise vector, $\epsilon\sim\mathcal{N}(0_{dim(S)},I_{dim(S)})$.}
  \item[{$\epsilon_{\theta}(s_{f_i}, i, s, a)$}] {Neural network noise predictor, 
    $S\times\{1,\dots,K\} \times S\times A\to \mathbb{R}^{\dim(s)}$.}
  \item[{$\alpha_{i},\bar{\alpha}_{i}$}] {Diffusion noising schedule coefficients in $(0,1)$.}
  \item[{$\beta_{i}$}] {Diffusion denoising schedule coefficient in $(0,1)$.}
  \item[$L_{\mathrm{DDPM}}$] DDPM denoising loss,
    \[
      L_{\mathrm{DDPM}}(\theta)
      =E_{x_{0},i,\epsilon}\bigl[\|\epsilon-\epsilon_{\theta}(x_{i},i)\|^{2}\bigr].
    \], or 
    \[
      L_{\mathrm{DDPM}}(\theta)
      =E_{s_f \sim \sr(\cdot |s, a),i,\epsilon}\bigl[\|\epsilon-\epsilon_{\theta}(s_{f_i},i, s, a)\|^{2}\bigr].
    \]
    when applied to the state successor measure
  \item[{$L_{\mathrm{BDM}}$}] {Bellman diffusion loss (see Sec.~4.1), enforcing Bellman flow on the DDPM objective.}
  \item[{$\sr_\tar$}] {Target‑network copy of the diffusion model density, parameterized by delayed weights.}
  \item[{$\tau$}] {Trajectory, $\tau=(s_{0},a_{0},s_{1},a_{1},\dots)$.}
  \item[$\mathcal{D}$] Offline dataset: multiset of trajectories drawn from the behavior policy.
  \item[$\mathcal{L}_{\mathrm{BC}}$] Behavior cloning loss,
    \[
      \mathcal{L}_{\mathrm{BC}}(\pi)
      =E_{(s,a)\sim\mathcal{D}}\bigl[-\log\pi(a\mid s)\bigr].
    \]
\end{description}

\section{Derivation}
   We derive a temporal difference (TD) update for diffusion models representing the successor state measure. Just as Q-learning enables off-policy training of the value function, this update rule makes it possible to learn the successor measure of arbitrary policies from a fixed data set. 

Our derivation consists of two main steps. First, we find an upper bound for the KL divergence between two diffusion models. This result may also be of independent interest. And second, we derive an update rule by  minimizing the KL divergence between the successor state measure and its Bellman update, analogous to minimizing the Bellman error in value learning.

    \subsection{KL Divergence between Diffusion Models}

    \begin{lemma}\label{lem: diff_kl}
        Let $q$ and $p$ be $K$-step diffusion models with noise schedule $\beta_i$, parameterized by neural networks with outputs $\epsilon_q$ and $\epsilon_p$, respectively. Let $q_i$ and $p_i$ be the distribution of the samples generated by the first $K-i$ steps of the forward process of $q$ and $p$, respectively. 
        Then 
        \begin{align*}
            KL&(q_0 || p_0) \leq \\& (K-1) E_{i \sim [1,K], x_i \sim q_i}\left[\frac{\beta_i||\epsilon_q(x_i, i) - \epsilon_p(x_i, i)||^2}{2 \alpha_i (1-\Bar{\alpha_i})}\right]
        \end{align*}
    \end{lemma}

    We provide a full proof in Appendix \ref{sec: KL divergence}, but provide a brief outline here for intuition. We can write the final output of the model as a sum of the contributions from each step. Using Jensen's inequality, we break this up into the sum of the KL divergences between the models at each step. Each diffusion step produces a Gaussian distribution, and the divergence between Gaussians with the same variance is the squared difference of their means (up to a constant). Adding these up gives us the expression above. 

    \subsection{Bounding the Bellman Flow Divergence}
    \begin{proposition}\label{prop: Bellman Flow Bound}
    Let $M$ be a Markov Decision Process with state space $S$, action space $A$, transition distribution $T$, reward function $R$, and discount rate $\gamma$.

    Let $KL_{Bellman}$ be defined as follows:
    \begin{align*}
        KL&_{Bellman} = \\&KL((1-\gamma) T(s' = s_f | s, a) + \\&\gamma E_{a' \sim \pi(s'),  s' \sim T( \cdot | s, a)}[d^{\pi}_\theta(s_f | s', a')] || d^{\pi}_\theta(s_f | s, a) ) 
    \end{align*}

    Then
    \begin{align*}
        KL&_{Bellman} \leq (K-1)E_{i, \epsilon, s' \sim T( \cdot | s, a)} [\frac{\beta_t}{2 \alpha_i (1-\Bar{\alpha_i})} \\&[
        (1-\gamma)|| \epsilon - \epsilon_\theta(s'_i, s, a, i) ||^2   \\&+ \gamma E_{a' \sim \pi(s'), s_f \sim d^\pi_\theta(\cdot | s, a)}\\&[|| \epsilon_\theta(s_{f_i}, s, a, i)  - \epsilon_\theta(s_{f_i}, s', a', i) ||^2 ]]] - H(T( \cdot | s, a))\\
    \end{align*}
    \end{proposition}

    The full proof of this result is in Appendix \ref{sec: Bellman Flow Bound}. As an overview, we use Jensen's inequality to separate the loss into the divergence between $T$ and $d^{\pi}_\theta$ and the divergence between $d^{\pi}_\theta( \cdot | s, a)$ and $d^{\pi}_\theta( \cdot | s', a')$. The first term can be bounded using the standard diffusion inequality presented in \citet{ho2020denoisingdiffusionprobabilisticmodels}, and the second term can be bounded using Lemma \ref{lem: diff_kl}. 

\subsection{TD Update}

    The above proposition finds an upper bound on the KL divergence between the left and right side of the Bellman constraints. We intend to learn an approximation of $\sr$ that minimizes this upper bound by semigradient TD.

    A straightforward application of semigradient TD would result in the following objective: 
    \begin{align*}
        KL&_{Bellman} \leq E_{i, \epsilon, s' \sim T( \cdot | s, a)} [\frac{1}{2 \alpha_i (1-\Bar{\alpha_i})} 
        \\& [(1-\gamma)|| \epsilon - \epsilon_\theta(s'_i, s, a, i) ||^2   
        \\&+ \gamma E_{a' \sim \pi(s'), s_f \sim d^\pi_\theta(\cdot | s', a')}[
        \\&|| \epsilon_\theta(s_{f_i}, s, a, i)  - \text{StopGrad(}\epsilon_\theta(s_{f_i}, s', a', i)) ||^2 ]]] 
        \\& - H(T( \cdot | s, a))\\
    \end{align*}
    However, attempting to learn with this objective directly would result in a method that was unnecessarily complicated, unstable, and poorly performing. We now make a number of changes based on common empirical practice to address these issues, and make the algorithm practical and amenable to realistic experiments. 
    

    First, we we observe that since $- H(T( \cdot | s, a)) $ does not depend on the parameters of the network, it will not affect the gradient and can be ignored. 

    Second, we note that the coefficient $\frac{\beta_i}{2 \alpha_i (1-\Bar{\alpha_i})}$ does not affect the fixed point of the gradient update, so it may be removed during training without changing the optimal solution. This is standard practice for diffusion models -- implementations rarely include this term, and removing it typically improves performance in practice \citep{ho2020denoisingdiffusionprobabilisticmodels}. 

    Finally, semigradient methods like TD are known to be unstable when combined with function approximation \citep{BairdCounterexample}. As a result, it is standard practice to replace the target value (in this case, StopGrad($\epsilon_\theta(s_{f_i}, s', a', i)$)) with a target network $\epsilon_\tar$ updated by exponential averaging of $\epsilon_\theta$. This is a widely accepted practice for deep reinforcement learning, and makes learning more stable both in theory and in practice \citep{chen2022targetnetworktruncationovercome, DQN, DDPG, TD3, SAC, PPO}. We will later show that these changes are consistent with minimizing the Bellman KL divergence, and that this loss has zero gradient when the Flow Constraints are satisfied. 
    
    \begin{equation}\label{eqn:BDM Loss}
    \begin{split}
        L_{BDM} = &E_{i, \epsilon, s' \sim T( \cdot | s, a)} [
        [
        (1-\gamma)\underbrace{|| \epsilon - \epsilon_\theta(s'_i, i) ||^2}_\text{Standard diffusion loss}   
        \\&+ \gamma E_{a' \sim \pi(s'), s_f \sim d^\pi_\tar(\cdot | s, a)} 
        \\&  [\underbrace{|| \epsilon_\theta(s_{f_i}, s, a, i)  - \epsilon_\tar(s_{f_i}, s', a', i) ||^2}_\text{Bellman backup loss}]]]
    \end{split}        
    \end{equation}
    
    The final loss contains two terms: a normal diffusion loss which attempts to make the network predict $s'$, and a Bellman consistency term that attempts to make states that are probable under $d^\pi_\theta(\cdot | s', a')$ also be probable under $d^\pi_\theta(\cdot | s, a)$

    This learning rule offers two main advantages compared to directly using the cross entropy from the future state distribution. Firstly, the bound from Lemma \ref{lem: diff_kl} is lower variance than directly estimating the cross entropy. In the standard DDPM loss, the target value for a given $x_i$ depends on the original unnoised value $x_0$. Since different values of $x_0$ can noise to the same $x_i$, this means the target values for $x_i$ are random. By contrast, the target values of the loss derived in Lemma \ref{lem: diff_kl} are deterministic and depend only on $x_i, s', a', $ and $i$. 
    If the transition function and policy are deterministic, this means the second term has zero variance. 
    Since we set $\gamma=0.99$ in our experiments, this means that we eliminate about 99\% of the target value variance from our training objective. This is effectively the same tradeoff made in using TD learning for the value function -- we exchange an unbiased, high-variance estimator for a potentially biased but low-variance estimator.  Given the success of TD learning in deep RL, we should expect this to be a very good tradeoff. 
    
    Second, the policy from which $a'$ is sampled does not need to be the same policy used to gather the dataset. This means we can train the Bellman Diffusion Model off-policy by using the target network to generate samples, and then using the above loss to learn to reproduce those samples.

    \subsection{Algorithm}

    We now present an algorithm for learning $d^\pi_\theta$.

    \begin{algorithm}[tbh]
      \caption{Calculate $d^{\pi}$ Loss}
      \label{alg:Train SSM}
    \begin{algorithmic}
    \STATE {\bfseries Have:} Networks $\epsilon_\theta$, $\epsilon_\tar$ number of diffusion steps $K$, policy $\pi$; 
    \STATE {\bfseries Input:} $(s, a, s')$
    \STATE Sample $a' \sim \pi(s')$
    \STATE Sample $i \sim Uniform([1, K])$
    \STATE Sample $\epsilon \sim \mathcal{N}(0,1)$
    \STATE $s_{f} \sim d^\pi_\tar(\cdot | s', a')$
    \STATE $s'_i = \sqrt{\Bar{\alpha_i}} s' + \sqrt{1-\Bar{\alpha_i}}\epsilon$
    \STATE $s_{f_i} = \sqrt{\Bar{\alpha_i}} s_f + \sqrt{1-\Bar{\alpha_i}}\epsilon$
    \STATE $L_1 = || \epsilon - \epsilon_\theta(s'_i, s, a, i)||^2$
    \STATE $L_2 = || \epsilon_{target}(s_{f_i}, s', a', i) - \epsilon_\theta(s_{f_i}, s, a, i)||^2$
    \STATE $L = (1-\gamma)L_1 + \gamma L_2$
    \RETURN $L$;
    \end{algorithmic}
    \end{algorithm}

    This is a simple change to the standard diffusion loss. Once we sample a state, action, next state tuple, we additionally find the next action $a'$, and use it to sample a future state $s_f$ from our target network. Then, we noise both the next state $s'$ and the future state $s_f$. We find the squared error for both, using the normal diffusion target $\epsilon$ for the network at $s'_i$ and the TD target $\epsilon_{target}(s_{f_i}, s', a', i)$ for  the network at $s'_{f_i}$. Finally, we weight the losses by $\gamma$ and $1-\gamma$ respectively, and return their sum. 

\section{Theoretical analysis}

A number of questions can now be asked about the proposed method, but perhaps the most pressing is whether this learns the correct distribution of future states. Since this method is related to Deep TD, full guarantees are not generally possible, as this family of methods can diverge in some cases \citep{BairdCounterexample}. However, it is possible to show that the optimal $\sr_\theta$ is a fixed point of the update operator, which is the same guarantee typically given for deep TD algorithms. We present the proof of this below.

Because we update $\epsilon_\theta$ by gradient descent, there is a fixed point when the gradient is zero. Thus, to show that the fixed point of a Bellman diffusion model is the same as the normal diffusion model, we begin with the normal diffusion model, and show that if the gradient of the standard diffusion loss is zero and $\sr_\theta = \sr_\tar = \sr$, then the gradient of the Bellman diffusion loss is also zero.

For convenience, we use the form of both losses that predicts $x_0$, rather than the version that predicts $\epsilon$. The proof is possible either way, but this form avoids needless calculation. The full proof is in the appendix, but this is a simple calculation based on applying the bias-variance decomposition to the loss. 

\begin{proposition}\label{prop:fixed_point}
    Let $\epsilon \sim \mathcal{N}(0,1)$ and $i \sim Unif(1,K)$.
    Let $L_{DDPM} = E_{x_0 \sim \sr(\cdot | s, a), \epsilon, i, (s, a) \sim D}[|| x_0- x_\theta(x_i, s', \pi(s'), i)||^2]$. 
    Suppose $\sr_\theta = \sr_{target} = \sr$ and $x_{target}(x_i, s, a, i) = E[x_0 | x_i, s, a, i]$. 

    Then $L_{DDPM} = L_{BDM} +  \gamma Var(x_0 | x_i, s', a', i)$
\end{proposition}

\begin{corollary}    
    $\grad_\theta L_{DDPM} = 0$ if and only $\grad_\theta L_{BDM} = 0$.
\end{corollary}

As we can see, the DDPM loss decomposes to the Bellman Diffusion loss, plus a variance term when $x_{target}(s, a, i) = E[x_0 | x_i, s_i, a_i, i]$. 
Additionally, the presence of the extra variance term strongly suggests that the BDM loss has a lower variance gradient. Intuitively, it is easy to see why this would the case -- for a given sample of $x_i$ (other than $x_0 = s'$), BDM has a deterministic target of $x_\tar$, while DDPM has the nondeterministic target $x_0$.

\section{Offline Reinforcement Learning Algorithm}

A central problem in offline RL is that, although it is possible to imitate the expert policy on the dataset, errors take the policy away from the dataset to new states where it cannot recover. Typically, it is difficult to control this future behavior because it is difficult to predict the future state occupancy of the agent. 

We propose a simple solution: rather than performing behavior cloning on the actions alone, we perform behavior cloning on the actions and future states. Intuitively, this ensures that the future distribution of states remains close to the dataset, minimizing distribution shift.  We do this by adding the cross entropy of the actual future state distribution and the Bellman Diffusion Model as a regularization term on the loss. Since the BDM depends on the action given by the policy, this loss can be backpropagated through the BDM to $\pi$. In Appendix \ref{sec: ergodic}, we show that if the environment is ergodic, this regularization scheme can be derived as an upper bound to the KL divergence between the data distribution and the policy's state-action occupancy measure. 

    We base our algorithm on ReBRAC, a state of the art method in offline reinforcement learning \citep{tarasov2023revisitingminimalistapproachoffline}. 
    We propose modifying ReBRAC to include this state imitation loss, in addition to the standard behavior cloning loss to encourage the policy to stay within the support of the dataset. We call this 
    algorithm TD3-SBC. This method can also be adapted for imitation learning by simply removing the value term and focusing on the imitation loss alone.

    \begin{algorithm}[tbh]
      \caption{TD3-SBC policy loss}
      \label{alg:Offline RL loss}
    \begin{algorithmic}
    \STATE {\bfseries Have:} 
    state space with dimension $d_S$, diffusion generation algorithm $G$, diffusion model network $\epsilon_\theta$, 
    policy network $\pi_\phi(a | s)$ with weights $\phi$
    Q-networks $Q_{\psi_1}$ and $Q_{\psi_2}$, 
    imitation loss weights $w_s$(for states) and $w_{a}$ (for actions)
    \STATE {\bfseries Input:} Training batch $\{(s, a, s')\}$
    
    \STATE \textit{\#Calculate action imitation loss}
    
    \STATE $L_a = -log(\pi_\phi(a | s))$
    
    \STATE\textit{\#Calculate state imitation loss}
    
    \STATE Sample future visited state $s_f$ from future trajectory of $s$
    \STATE Sample $i \sim Uniform([1, K])$, where $K$ is the number of diffusion steps
    \STATE Sample $\epsilon \sim \mathcal{N}(0,I_{d_S})$
    \STATE $s_{f_i} \leftarrow \sqrt{\Bar{\alpha_i}} s_f + \sqrt{1-\Bar{\alpha_i}}\epsilon$
    \STATE $a_\pi \sim \pi_\phi(\cdot | s)$
    \STATE $\eta_i = \frac{\beta_i}{\alpha_i(1-\alpha_i)}$
    \STATE $L_s \leftarrow  \eta_i || \epsilon - \epsilon_\theta(s_{f_i}, s, a_\pi, i)||^2$
    
    \STATE\textit{\#Calculate value loss}
    \STATE $L_Q \leftarrow \min{(Q_{\psi_1}(s, a_\pi), Q_{\psi_2}(s, a_\pi))}$
    \RETURN $L = L_Q + w_{s}L_s +  w_{a}L_a$;
    \end{algorithmic}
    \end{algorithm}

While other methods use a similar objective function, either in imitation learning or in offline reinforcement learning, it has not previously been possible to directly optimize this objective. Instead, other methods take the convex or Fenchel dual of the problem and optimize that, which typically results in a point-reweighting objective ~\citep{ho2016generativeadversarialimitationlearning, ma2022versatileofflineimitationobservations, nachum2020reinforcementlearningfenchelrockafellarduality, lee2021optidiceofflinepolicyoptimization, ma2022farillgooffline}. Unfortunately, many methods derived by this method, such as the DICE family of methods, fail to achieve results competitive with standard actor critic methods \citep{park2024valuelearningreallymain}. 

Our method allows us to regularize the state distribution in offline RL while still actor-critic methods. We include the full TD3-SBC algorithm below. 

\begin{algorithm}[tbh]
      \caption{TD3-SBC algorithm}
      \label{alg:full learning alg}
    \begin{algorithmic}
    \STATE {\bfseries Have:} 
    state space with dimension $d_S$, 
    diffusion generation algorithm $G$, 
    diffusion model network $\epsilon_\theta$, 
    policy network $\pi_\phi(a | s)$ with weights $\phi$
    Q-networks $Q_{\psi_1}$ and $Q_{\psi_2}$, 
      
    Randomly initialize critic networks Q-networks $Q_{\psi_1}$ and $Q_{\psi_2}$ with weights $\psi_1$ and $\psi_2$, diffusion network $\epsilon_\theta$ with weights $\theta$, policy network $\pi_\phi(a | s)$ with weights $\phi$. 
    Initialize target networks $Q_{\tar_1}$, $Q_{\tar_2}$, and $\epsilon_\tar$ with copies of the weights from $Q_{\psi_1}$ $Q_{\psi_2}$, and $\epsilon_\theta$, respectively.

    \STATE {\bfseries Input:} imitation loss weights $w_s$(for states) and $w_{a}$ (for actions), polyak averaging coefficient $\tau$, action noise $\sigma_a^2$

    \FOR{episode = 1, M}
    \STATE Sample a batch $B$ of $(s, a, r, s')$ transitions from the dataset
    \STATE \textit{\#Critic update}
    \STATE Sample $a' \sim \mathcal{N}(\pi_\phi(s), \sigma_a^2)$
    \STATE {\color{blue} Update $\sr_\theta$, as described in Algorithm \ref{alg:Train SSM}}
    \STATE $y \leftarrow r + \gamma \min(Q_{\tar_1}(s', a'), Q_{\tar_2}(s', a'))$
    \STATE Update $Q_{\psi_1}, Q_{\psi_2}$ by minimizing the Q loss $\mathcal{L}_Q = ||Q_{\psi_1} - y||^2 + ||Q_{\psi_2} - y||^2$
    \STATE \textit{\#Policy update}
    \STATE {\color{blue} Calculate the policy loss $\mathcal{L}_\pi$ as described in Algorithm \ref{alg:Offline RL loss}}
    \STATE Update the policy by minimizing $\mathcal{L}_\pi$
    \ENDFOR
    \end{algorithmic}
    \end{algorithm}

Blue text represents additions to TD3-BC. 

\section{Experiments}

\begin{table*}[t]
\caption{D4RL Performance} \label{tab:d4rl-performance}

\begin{center}
\resizebox{\textwidth}{!}{
\begin{tabular}{lrrrrrr}
\textbf{Task Name} & \textbf{TD3+BC} & \textbf{IQL} & \textbf{CQL} & \textbf{SAC-RND} & \textbf{ReBRAC} & \textbf{TD3-SBC} \\
\hline \\

halfcheetah-medium         & 54.7 $\pm$ 0.9   & 50.0 $\pm$ 0.2   & 46.9 $\pm$ 0.4   & \textbf{66.4 $\pm$ 1.4}   & 65.6 $\pm$ 1.0   & 65.0 $\pm$ 0.8 \\
halfcheetah-expert         & 93.4 $\pm$ 0.4   & 95.5 $\pm$ 2.1   & 97.3 $\pm$ 1.1   & 102.6 $\pm$ 4.2   & \textbf{105.9 $\pm$ 1.7}   & \textbf{105.9 $\pm$ 0.7} \\
halfcheetah-medium-expert  & 89.1 $\pm$ 5.6   & 92.7 $\pm$ 2.8   & 95.0 $\pm$ 1.4   & \textbf{108.1 $\pm$ 1.5}   & 101.1 $\pm$ 5.2   & 105.8 $\pm$ 1.8 \\
halfcheetah-medium-replay  & 45.0 $\pm$ 1.1   & 42.1 $\pm$ 3.6   & 45.3 $\pm$ 0.3   & 51.2 $\pm$ 3.2   & 51.0 $\pm$ 0.8   & \textbf{54.7 $\pm$ 0.5} \\
\hline \\

hopper-medium              & 60.9 $\pm$ 7.6   & 65.2 $\pm$ 4.2   & 61.9 $\pm$ 6.4   & 91.1 $\pm$ 10.1   & \textbf{102.0 $\pm$ 1.0}   & 101.6 $\pm$ 0.3 \\
hopper-expert              & 109.6 $\pm$ 3.7  & 108.8 $\pm$ 3.1  & 106.5 $\pm$ 9.1  & 109.8 $\pm$ 0.5   & 100.1 $\pm$ 8.3   & \textbf{111.5 $\pm$ 0.7} \\
hopper-medium-expert       & 87.8 $\pm$ 10.5  & 85.5 $\pm$ 29.7  & 96.9 $\pm$ 15.1  & \textbf{109.8 $\pm$ 0.6}   & 107.0 $\pm$ 6.4   & 107.3 $\pm$ 2.7 \\
hopper-medium-replay       & 55.1 $\pm$ 31.7  & 89.6 $\pm$ 13.2  & 86.3 $\pm$ 7.3   & 97.2 $\pm$ 9.0   & 98.1 $\pm$ 5.3   & \textbf{101.1 $\pm$ 1.2} \\
\hline \\

walker2d-medium            & 77.7 $\pm$ 2.9   & 80.7 $\pm$ 3.4   & 79.5 $\pm$ 3.2   & \textbf{92.7 $\pm$ 1.2}   & 82.5 $\pm$ 3.6   & 88.7 $\pm$ 1.5 \\
walker2d-expert            & 110.0 $\pm$ 0.6  & 96.9 $\pm$ 32.3  & 109.3 $\pm$ 0.1  & 104.5 $\pm$ 22.8   & \textbf{112.3 $\pm$ 0.2}   & 111.8 $\pm$ 0.1 \\
walker2d-medium-expert     & 110.4 $\pm$ 0.6  & 112.1 $\pm$ 0.5  & 109.1 $\pm$ 0.2  & 104.6 $\pm$ 11.2   & \textbf{111.6 $\pm$ 0.3}   & 110.7 $\pm$ 0.3 \\
walker2d-medium-replay     & 68.0 $\pm$ 19.2  & 75.4 $\pm$ 9.3   & 76.8 $\pm$ 10.0  & \textbf{89.4 $\pm$ 3.8}   & 77.3 $\pm$ 7.9   & 87.9 $\pm$ 3.0 \\
\hline \\

Average                    & 80.1 & 82.9 & 84.2 & 94.0 & 92.9 & \textbf{96.0} \\
\hline
\end{tabular}
}
\end{center}

\caption{D4RL performance. Reported values are the last-iterate return averaged across all training seeds. The $\pm$ symbol represents the standard deviation across seeds. All results other than TD3-SBC are taken from ~\citet{tarasov2023revisitingminimalistapproachoffline}}.

\end{table*}

To evaluate TD3-SBC, we compare its performance against ReBRAC with the standard behavior cloning loss. We focus on three D4RL tasks: hopper, halfcheetah, and walker2d \citep{fu2021d4rldatasetsdeepdatadriven}. We evaluate all methods on four datasets for each task. 
\begin{enumerate}
    \item One million data points from a partially-trained RL agent (medium)
    \item The full medium dataset as well as one million datapoints from the agent's replay buffer (medium-replay)
    \item One million datapoints from a fully-trained RL agent (expert)
    \item Union of the medium and expert datasets (medium-expert)
\end{enumerate}

All methods are trained for $10^6 $ steps. For the most part, hyperparameters are taken from ReBRAC. However, the action behavior cloning weight, the state behavior cloning weight, and the actor and diffusion model learning rates were tuned. Table \ref{tab:d4rl-performance} reports the results of these experiments. We found that for some environments, adding state behavior cloning allowed us to reduce the action behavior cloning weight, giving the model greater freedom to pursue high rewards. Additionally, we found that the state behavior cloning loss could still be too high variance for the policy, so we experimented with decreasing the actor learning rate when this led to instability.

We find that TD3-SBC outperforms all baselines on three environments (halfcheetah-medium-replay, hopper-expert, and hopper-medium-replay) and ties  with another method for a fourth (halfcheetah-expert). For another five environments, TD3-SBC has the second highest expected reward of the group. As a result, TD3-SBC has the highest average reward of the methods. Additionally, TD3-SBC's reward has a substantially lower variance, which may indicate greater stability as a result of better regularization.





\section{Conclusion}

Optimization of the state occupancy measure and successor state measure are topics of great interest to offline reinforcement learning research, because they offer a valuable theoretical framework for controlling distribution shift. However, methods based on this approach have typically been held back by the difficulty of estimating these distributions. We propose a low-variance off-policy TD-based learning algorithm for estimating the state successor measure, and show that it can be used to regularize existing offline RL algorithms. This makes it possible to solve the primal problem by directly optimizing the state occupancy measure, instead of resorting to the Fenchel dual problem as many other methods do. This approach opens up a number of new possibilities for both online and offline reinforcement learning, as state occupancy formulations can more easily be proposed, experimented with, and combined with existing RL methods. Unlike GAIL or the DICE family of algorithms, the method we propose can be used, modified, and extended without lengthy derivation or extensive knowledge of convex analysis techniques like convex duality \citep{ho2016generativeadversarialimitationlearning, nachum2019algaedicepolicygradientarbitrary, ma2022versatileofflineimitationobservations}. Moreover, we find that the proposed algorithm has strong performance on the D4RL offline reinforcement learning dataset, and achieves new records on multiple environments.

\bibliography{bib} 

\begin{thebibliography}{}

\bibitem[Chen et~al., 2022a]{chen2022learninginfinitehorizonaveragerewardmarkov}
Chen, L., Jain, R., and Luo, H. (2022a).
\newblock Learning infinite-horizon average-reward markov decision processes with constraints.

\bibitem[Chen et~al., 2022b]{chen2022targetnetworktruncationovercome}
Chen, Z., Clarke, J.~P., and Maguluri, S.~T. (2022b).
\newblock Target network and truncation overcome the deadly triad in $q$-learning.

\bibitem[Chi et~al., 2023]{chi2023diffusionpolicy}
Chi, C., Feng, S., Du, Y., Xu, Z., Cousineau, E., Burchfiel, B., and Song, S. (2023).
\newblock Diffusion policy: Visuomotor policy learning via action diffusion.
\newblock In {\em Proceedings of Robotics: Science and Systems (RSS)}.

\bibitem[Dann et~al., 2023]{dann2023bestworldspolicyoptimization}
Dann, C., Wei, C.-Y., and Zimmert, J. (2023).
\newblock Best of both worlds policy optimization.

\bibitem[Dayan, 1993]{dayan1993successorrepresentations}
Dayan, P. (1993).
\newblock Improving generalization for temporal difference learning: The successor representation.
\newblock {\em Neural Computation}, 5(4):613--624.

\bibitem[Dhariwal and Nichol, 2021]{dhariwal2021diffusionmodelsbeatgans}
Dhariwal, P. and Nichol, A. (2021).
\newblock Diffusion models beat gans on image synthesis.

\bibitem[Farebrother et~al., 2025]{farebrother2025temporaldifferenceflows}
Farebrother, J., Pirotta, M., Tirinzoni, A., Munos, R., Lazaric, A., and Touati, A. (2025).
\newblock Temporal difference flows.

\bibitem[Fu et~al., 2021]{fu2021d4rldatasetsdeepdatadriven}
Fu, J., Kumar, A., Nachum, O., Tucker, G., and Levine, S. (2021).
\newblock D4rl: Datasets for deep data-driven reinforcement learning.

\bibitem[Fujimoto et~al., 2018]{TD3}
Fujimoto, S., van Hoof, H., and Meger, D. (2018).
\newblock Addressing function approximation error in actor-critic methods.

\bibitem[Haarnoja et~al., 2018]{SAC}
Haarnoja, T., Zhou, A., Abbeel, P., and Levine, S. (2018).
\newblock Soft actor-critic: Off-policy maximum entropy deep reinforcement learning with a stochastic actor.
\newblock In Dy, J. and Krause, A., editors, {\em Proceedings of the 35th International Conference on Machine Learning}, volume~80 of {\em Proceedings of Machine Learning Research}, pages 1861--1870. PMLR.

\bibitem[Ho and Ermon, 2016]{ho2016generativeadversarialimitationlearning}
Ho, J. and Ermon, S. (2016).
\newblock Generative adversarial imitation learning.

\bibitem[Ho et~al., 2020]{ho2020denoisingdiffusionprobabilisticmodels}
Ho, J., Jain, A., and Abbeel, P. (2020).
\newblock Denoising diffusion probabilistic models.

\bibitem[Janner et~al., 2022]{janner2022planningdiffusionflexiblebehavior}
Janner, M., Du, Y., Tenenbaum, J.~B., and Levine, S. (2022).
\newblock Planning with diffusion for flexible behavior synthesis.

\bibitem[Janner et~al., 2021]{janner2021generativetemporaldifferencelearning}
Janner, M., Mordatch, I., and Levine, S. (2021).
\newblock Generative temporal difference learning for infinite-horizon prediction.

\bibitem[Jin et~al., 2023]{jin2023improvedbestofbothworldsguaranteesmultiarmed}
Jin, T., Liu, J., and Luo, H. (2023).
\newblock Improved best-of-both-worlds guarantees for multi-armed bandits: Ftrl with general regularizers and multiple optimal arms.

\bibitem[Lee et~al., 2021]{lee2021optidiceofflinepolicyoptimization}
Lee, J., Jeon, W., Lee, B.-J., Pineau, J., and Kim, K.-E. (2021).
\newblock Optidice: Offline policy optimization via stationary distribution correction estimation.

\bibitem[Lee et~al., 2020]{lee2020efficientexplorationstatemarginal}
Lee, L., Eysenbach, B., Parisotto, E., Xing, E., Levine, S., and Salakhutdinov, R. (2020).
\newblock Efficient exploration via state marginal matching.

\bibitem[Lillicrap et~al., 2019]{DDPG}
Lillicrap, T.~P., Hunt, J.~J., Pritzel, A., Heess, N., Erez, T., Tassa, Y., Silver, D., and Wierstra, D. (2019).
\newblock Continuous control with deep reinforcement learning.

\bibitem[Ma et~al., 2022a]{ma2022versatileofflineimitationobservations}
Ma, Y.~J., Shen, A., Jayaraman, D., and Bastani, O. (2022a).
\newblock Versatile offline imitation from observations and examples via regularized state-occupancy matching.

\bibitem[Ma et~al., 2022b]{ma2022farillgooffline}
Ma, Y.~J., Yan, J., Jayaraman, D., and Bastani, O. (2022b).
\newblock How far i'll go: Offline goal-conditioned reinforcement learning via $f$-advantage regression.

\bibitem[Mnih et~al., 2013]{DQN}
Mnih, V., Kavukcuoglu, K., Silver, D., Graves, A., Antonoglou, I., Wierstra, D., and Riedmiller, M. (2013).
\newblock Playing atari with deep reinforcement learning.

\bibitem[Mutti et~al., 2021]{mutti2021taskagnosticexplorationpolicygradient}
Mutti, M., Pratissoli, L., and Restelli, M. (2021).
\newblock Task-agnostic exploration via policy gradient of a non-parametric state entropy estimate.

\bibitem[Nachum and Dai, 2020]{nachum2020reinforcementlearningfenchelrockafellarduality}
Nachum, O. and Dai, B. (2020).
\newblock Reinforcement learning via fenchel-rockafellar duality.

\bibitem[Nachum et~al., 2019]{nachum2019algaedicepolicygradientarbitrary}
Nachum, O., Dai, B., Kostrikov, I., Chow, Y., Li, L., and Schuurmans, D. (2019).
\newblock Algaedice: Policy gradient from arbitrary experience.

\bibitem[Park et~al., 2024]{park2024valuelearningreallymain}
Park, S., Frans, K., Levine, S., and Kumar, A. (2024).
\newblock Is value learning really the main bottleneck in offline rl?

\bibitem[Santi et~al., 2025]{desanti2025provablemaximumentropymanifold}
Santi, R.~D., Vlastelica, M., Hsieh, Y.-P., Shen, Z., He, N., and Krause, A. (2025).
\newblock Provable maximum entropy manifold exploration via diffusion models.

\bibitem[Schulman et~al., 2017]{PPO}
Schulman, J., Wolski, F., Dhariwal, P., Radford, A., and Klimov, O. (2017).
\newblock Proximal policy optimization algorithms.

\bibitem[Sun, 2023]{offinerlkit}
Sun, Y. (2023).
\newblock Offlinerl-kit: An elegant pytorch offline reinforcement learning library.
\newblock \url{https://github.com/yihaosun1124/OfflineRL-Kit}.

\bibitem[Tarasov et~al., 2023]{tarasov2023revisitingminimalistapproachoffline}
Tarasov, D., Kurenkov, V., Nikulin, A., and Kolesnikov, S. (2023).
\newblock Revisiting the minimalist approach to offline reinforcement learning.

\bibitem[Wang et~al., 2023]{wang2023diffusionpoliciesexpressivepolicy}
Wang, Z., Hunt, J.~J., and Zhou, M. (2023).
\newblock Diffusion policies as an expressive policy class for offline reinforcement learning.

\bibitem[Williams and Baird, 1993]{BairdCounterexample}
Williams, R.~J. and Baird, L.~C. (1993).
\newblock Analysis of some incremental variants of policy iteration: First steps toward understanding actor-cr.

\bibitem[Zhang et~al., 2020]{zhang2020variationalpolicygradientmethod}
Zhang, J., Koppel, A., Bedi, A.~S., Szepesvari, C., and Wang, M. (2020).
\newblock Variational policy gradient method for reinforcement learning with general utilities.

\bibitem[Zimmert and Seldin, 2022]{zimmert2022tsallisinfoptimalalgorithmstochastic}
Zimmert, J. and Seldin, Y. (2022).
\newblock Tsallis-inf: An optimal algorithm for stochastic and adversarial bandits.

\end{thebibliography}
\bibliographystyle{apalike}

\begin{enumerate}

  \item For all models and algorithms presented, check if you include:
  \begin{enumerate}
    \item A clear description of the mathematical setting, assumptions, algorithm, and/or model. Yes
    \item An analysis of the properties and complexity (time, space, sample size) of any algorithm. No -- we do not analyze sample complexity or convergence rate, and the algorithm simplely loops until convergence
    \item (Optional) Anonymized source code, with specification of all dependencies, including external libraries. Yes -- this will be included in the supplementary materials
  \end{enumerate}

  \item For any theoretical claim, check if you include:
  \begin{enumerate}
    \item Statements of the full set of assumptions of all theoretical results. Yes
    \item Complete proofs of all theoretical results. Yes
    \item Clear explanations of any assumptions. Yes
  \end{enumerate}

  \item For all figures and tables that present empirical results, check if you include:
  \begin{enumerate}
    \item The code, data, and instructions needed to reproduce the main experimental results (either in the supplemental material or as a URL). Yes
    \item All the training details (e.g., data splits, hyperparameters, how they were chosen). Yes
    \item A clear definition of the specific measure or statistics and error bars (e.g., with respect to the random seed after running experiments multiple times). Yes
    \item A description of the computing infrastructure used. (e.g., type of GPUs, internal cluster, or cloud provider). Yes
  \end{enumerate}

  \item If you are using existing assets (e.g., code, data, models) or curating/releasing new assets, check if you include:
  \begin{enumerate}
    \item Citations of the creator If your work uses existing assets. Yes
    \item The license information of the assets, if applicable. Not Applicable
    \item New assets either in the supplemental material or as a URL, if applicable. Not Applicable
    \item Information about consent from data providers/curators. Not Applicable
    \item Discussion of sensible content if applicable, e.g., personally identifiable information or offensive content. Not Applicable
  \end{enumerate}

  \item If you used crowdsourcing or conducted research with human subjects, check if you include:
  \begin{enumerate}
    \item The full text of instructions given to participants and screenshots. Not Applicable
    \item Descriptions of potential participant risks, with links to Institutional Review Board (IRB) approvals if applicable. Not Applicable
    \item The estimated hourly wage paid to participants and the total amount spent on participant compensation. Not Applicable
  \end{enumerate}

\end{enumerate}

\clearpage
\appendix
\thispagestyle{empty}

\onecolumn
\aistatstitle{Appendix}


\section{Proofs mentioned in the main text}

\subsection{KL divergence between diffusion models}
\label{sec: KL divergence}

\textbf{Lemma 1.}\label{lem:diff_kl_revised}
Let $q$ and $p$ be $K$-step discrete-time diffusion models with forward noising kernels
\[
q(x_i\mid x_{i-1})=\mathcal{N}(\sqrt{\alpha_i}x_{i-1},\beta_i I),
\qquad \alpha_i = 1-\beta_i,
\qquad \bar\alpha_i=\prod_{j=1}^i \alpha_j,
\]
and suppose both reverse conditionals $q(x_{i-1}\mid x_i)$ and $p(x_{i-1}\mid x_i)$ are Gaussian with {\bf the same} variance $\beta_i I$ (this is the common DDPM convention). Let the corresponding noise-predicting networks be denoted by $\epsilon_q(x_i,i)$ and $\epsilon_p(x_i,i)$, and let the reverse-model means take the DDPM parameterized form
\[
\mu_\theta(x_i,i)=\frac{1}{\sqrt{\alpha_i}}\Big(x_i - \frac{\beta_i}{\sqrt{1-\bar\alpha_i}}\,\epsilon_\theta(x_i,i)\Big).
\]

Then
\[
\boxed{%
KL(q_0 \,\|\, p_0) \;\le\; (K-1)\,\mathbb{E}_{i\sim\mathrm{Unif}\{1,\dots,K\},\,x_i\sim q_i}
\left[\frac{\beta_i}{2\alpha_i(1-\bar\alpha_i)}\|\epsilon_q(x_i,i)-\epsilon_p(x_i,i)\|^2\right].}
\]

\begin{proof}
By assumption the reverse-conditionals of both models are Gaussian with equal covariance matrix $\beta_i I$; this lets us use the closed form of KL between Gaussians with equal covariance.

Start with the chain-rule expansion and Jensen's inequality (or simply note KL is nonincreasing under marginalization):
\[
KL(q_0\|p_0) \le KL(q_{0:K}\|p_{0:K})
= \sum_{i=1}^{K} \mathbb{E}_{x_i\sim q_i}\big[ KL(q(x_{i-1}\mid x_i)\| p(x_{i-1}\mid x_i)) \big].
\]
For each fixed $i$, the two conditionals are Gaussians with the same covariance $\beta_i I$, hence
\[
KL\big(\mathcal{N}(\mu_q,\beta_i I)\,\big\|\,\mathcal{N}(\mu_p,\beta_i I)\big)
=\frac{1}{2\beta_i}\|\mu_q-\mu_p\|^2.
\]
Using the DDPM parameterization of the means,
\begin{align*}
\mu_q(x_i,i)-\mu_p(x_i,i)
&= \frac{1}{\sqrt{\alpha_i}}\Big( \frac{\beta_i}{\sqrt{1-\bar\alpha_i}}(\epsilon_p(x_i,i)-\epsilon_q(x_i,i)) \Big)\\
&= -\,\frac{\beta_i}{\sqrt{\alpha_i(1-\bar\alpha_i)}} \big(\epsilon_q(x_i,i)-\epsilon_p(x_i,i)\big).
\end{align*}
Substituting into the Gaussian-KL expression gives
\[
KL(q(x_{i-1}\mid x_i)\| p(x_{i-1}\mid x_i))
=\frac{1}{2\beta_i}\cdot\frac{\beta_i^2}{\alpha_i(1-\bar\alpha_i)}\|\epsilon_q-\epsilon_p\|^2
=\frac{\beta_i}{2\alpha_i(1-\bar\alpha_i)}\|\epsilon_q-\epsilon_p\|^2.
\]
Averaging over $i$ and $x_i\sim q_i$ and factoring $(K-1)$ yields the stated bound.
\end{proof}


\vspace{1em}

\subsection{Bellman Flow Bound}
\label{sec: Bellman Flow Bound}

\textbf{Proposition 1.}\label{prop:bellman_flow_revised}
Let $M$ be an MDP with state space $S$, action space $A$, transition kernel $T$, and discount factor $\gamma\in[0,1)$. Let $d^\pi_\theta(\cdot\mid s,a)$ denote a family of successor-state diffusion models (one diffusion model per conditioning $(s,a)$) that satisfy the assumptions of Lemma \ref{lem:diff_kl_revised} with a shared noise schedule $\{\beta_i\}$. Define
\[
\begin{aligned}
KL_{\mathrm{Bellman}} :=\:& KL\Big( (1-\gamma)T(\cdot\mid s,a)
+\gamma\,\mathbb{E}_{s'\sim T(\cdot\mid s,a),\,a'\sim\pi(\cdot\mid s')}[ d^\pi_\theta(\cdot\mid s',a') ] \big\|\, d^\pi_\theta(\cdot\mid s,a)\Big).
\end{aligned}
\]

Then
\[
\begin{aligned}
KL_{\mathrm{Bellman}} \le\; & (K-1)\,\mathbb{E}_{i\!,\epsilon\!,\,s'\sim T(\cdot\mid s,a)}\Bigg[\frac{\beta_i}{2\alpha_i(1-\bar\alpha_i)}\Big(
(1-\gamma)\|\epsilon - \epsilon_\theta(s'_i,s,a,i)\|^2 \\
&\qquad\qquad\qquad\qquad\qquad\qquad
+\gamma\,\mathbb{E}_{a'\sim\pi(s'),\,s_f\sim d^\pi_\theta(\cdot\mid s',a')}
\|\epsilon_\theta(s_{f_i},s,a,i)-\epsilon_\theta(s_{f_i},s',a',i)\|^2\Big)\Bigg]\\
&\qquad\qquad\qquad\qquad\qquad\qquad - H\big(T(\cdot\mid s,a)\big).
\end{aligned}
\]

\begin{proof}
 Because KL is convex in its first (left) argument,
\[
\begin{aligned}
KL_{\mathrm{Bellman}}
&= KL\big((1-\gamma)T(\cdot\mid s,a) + \gamma \mathbb{E}_{s',a'}[d^\pi_\theta(\cdot\mid s',a')] \,\|\, d^\pi_\theta(\cdot\mid s,a)\big) \\
&\le (1-\gamma)\,KL(T(\cdot\mid s,a)\|d^\pi_\theta(\cdot\mid s,a)) + \gamma\,KL(\mathbb{E}_{s',a'}[d^\pi_\theta(\cdot\mid s',a')]\|d^\pi_\theta(\cdot\mid s,a)),
\end{aligned}
\]
Next, by convexity of KL under mixture,
\[
KL\big(\mathbb{E}_{s',a'}[d^\pi_\theta(\cdot\mid s',a')]\big\| d^\pi_\theta(\cdot\mid s,a)\big)
\le \mathbb{E}_{s',a'}\big[ KL(d^\pi_\theta(\cdot\mid s',a') \| d^\pi_\theta(\cdot\mid s,a) )\big].
\]
Combining these two inequalities yields
\[
\begin{aligned}
KL_{\mathrm{Bellman}} \le\; &(1-\gamma)\,KL\big(T(\cdot\mid s,a)\| d^\pi_\theta(\cdot\mid s,a)\big) \\
&\qquad + \gamma\,\mathbb{E}_{s'\sim T(\cdot\mid s,a),\,a'\sim\pi(\cdot\mid s')}\big[ KL(d^\pi_\theta(\cdot\mid s',a') \| d^\pi_\theta(\cdot\mid s,a))\big].
\end{aligned}
\]
Using the identity $KL(q\|p) = -\mathbb{E}_{x\sim q}[\log p(x)] - H(q)$ on the first term gives
\[
(1-\gamma)\,KL(T\|d^\pi_\theta)= (1-\gamma)\,\mathbb{E}_{s'\sim T}[ -\log d^\pi_\theta(s'\mid s,a)] - (1-\gamma)H(T).
\]
Applying the DDPM-style ELBO inequality (see \citet{ho2020denoisingdiffusionprobabilisticmodels}) to bound $\mathbb{E}_{s'\sim T}[-\log d^\pi_\theta(s'\mid s,a)]$ yields
\[
\mathbb{E}_{s'\sim T}[ -\log d^\pi_\theta(s'\mid s,a)]
\le (K-1)\,\mathbb{E}_{i,\epsilon,s'\sim T}\left[\frac{\beta_i}{2\alpha_i(1-\bar\alpha_i)}\|\epsilon-\epsilon_\theta(s'_i,s,a,i)\|^2\right].
\]
For the second term, apply Lemma \ref{lem:diff_kl_revised} to the pair of diffusion models $d^\pi_\theta(\cdot\mid s',a')$ (left) and $d^\pi_\theta(\cdot\mid s,a)$ (right). This yields the inner expectation over $s_f\sim d^\pi_\theta(\cdot\mid s',a')$ and corresponding noisy $s_{f_i}$ samples:
\[
\begin{aligned}
&KL(d^\pi_\theta(\cdot\mid s',a') \| d^\pi_\theta(\cdot\mid s,a)) \\
&\qquad\le (K-1)\,\mathbb{E}_{i,\epsilon,s_f\sim d^\pi_\theta(\cdot\mid s',a')}
\left[\frac{\beta_i}{2\alpha_i(1-\bar\alpha_i)} \|\epsilon_\theta(s_{f_i},s,a,i)-\epsilon_\theta(s_{f_i},s',a',i)\|^2\right].
\end{aligned}
\]
Combining the two bounds (and collecting the entropy term $-H(T)$ once) gives the stated inequality.
\end{proof}

\vspace{1em}

\subsection{Fixed point of the update rule}
\label{sec:fixed_point_revised}

\textbf{Proposition 2.}\label{prop:fixed_point_revised}
Let $i\sim\mathrm{Unif}\{1,\dots,K\}$ and $\epsilon\sim\mathcal{N}(0,I)$. Define the DDPM $x_0$-prediction loss
\[
L_{\mathrm{DDPM}} \;=\; \mathbb{E}_{(s,a)\sim D}\,\mathbb{E}_{x_0\sim d^\pi(\cdot\mid s,a),\,i,\epsilon}\big[ \|x_0 - x_\theta(x_i,s,a,i)\|^2 \big],
\]
where $x_i=\sqrt{\bar\alpha_i}x_0+\sqrt{1-\bar\alpha_i}\epsilon$ and $x_\theta(\cdot)$ denotes the model's $x_0$-prediction. Suppose the learned successor models satisfy $d^\pi_\theta = d^\pi_{\mathrm{tar}} = d^\pi$ and define
\[
x_{\mathrm{target}}(x_i,s',a',i) := \mathbb{E}[x_0 \mid x_i, s', a', i].
\]
Then
\[
\boxed{ \; L_{\mathrm{DDPM}} \;=\; L_{\mathrm{BDM}} \;+\; \gamma\,\mathbb{E}_{x_i,s',a', i}\big[ \mathrm{Var}(x_0\mid x_i,s',a',i)\big] \;},
\]
where $L_{\mathrm{BDM}}$ is the Bellman-diffusion MSE term that replaces the stochastic target $x_0$ with the conditional mean $x_{\mathrm{target}}$ in the second (bootstrap) component of the mixture.

Consequently, under the stated assumptions the parameter values for which $\nabla_\theta L_{\mathrm{DDPM}}=0$ coincide with those for which $\nabla_\theta L_{\mathrm{BDM}}=0$.

\begin{proof}
Write the DDPM loss by expanding the successor mixture (Bellman flow decomposition of $d^\pi$):
\[
\begin{aligned}
L_{\mathrm{DDPM}}(s,a)
&= (1-\gamma)\,\mathbb{E}_{x_0\sim T(\cdot\mid s,a),\,i,\epsilon}\big[ \|x_0 - x_\theta\|^2\big] \\
&\quad + \gamma\,\mathbb{E}_{s'\sim T(\cdot\mid s,a),\,a'\sim\pi(\cdot\mid s'),\,x_0\sim d^\pi(\cdot\mid s',a'),\,i,\epsilon}\big[ \|x_0 - x_\theta\|^2\big].
\end{aligned}
\]
In the second term apply the law of total expectation conditioning on $(x_i,s',a',i)$:
\[
\begin{aligned}
&\mathbb{E}_{x_0\sim d^\pi(\cdot\mid s',a')}\big[ \|x_0 - x_\theta(x_i,s,a,i)\|^2 \mid x_i,s',a',i\big] \\
&\qquad = \| \mathbb{E}[x_0\mid x_i,s',a',i] - x_\theta(x_i,s,a,i)\|^2 + \mathrm{Var}(x_0\mid x_i,s',a',i).
\end{aligned}
\]
Replacing the inner conditional expectation by $x_{\mathrm{target}}$ yields
\[
\begin{aligned}
L_{\mathrm{DDPM}}(s,a)
&= (1-\gamma)\,\mathbb{E}_{T, x_i,i}[\|x_0 - x_\theta(x_i,s,a,i)\|^2] \\
&\quad + \gamma\,\mathbb{E}_{x_i,s',a', i}\big[ \|x_{\mathrm{target}}(x_i,s',a',i) - x_\theta(x_i,s,a,i)\|^2 + \mathrm{Var}(x_0\mid x_i,s',a',i)\big] \\
&= L_{\mathrm{BDM}}(s,a) + \gamma\,\mathbb{E}_{x_i,s',a', i}\big[ \mathrm{Var}(x_0\mid x_i,s',a',i)\big].
\end{aligned}
\]
The variance term is independent of the model parameters when $x_{\mathrm{target}}$ is the conditional mean; hence both losses share the same gradient zeros under the stated assumptions.
\end{proof}


\subsection{Connection to divergence from data distribution} \label{sec: ergodic}

One contribution of this work is bridging the gap between traditional policy gradient methods and DICE methods, which use a state occupancy objective. Here, we explore how our SSM regularization term can be seen as an upper bound on the divergence between the dataset and the state occupancy. 

To do this, we begin by formally defining the state-action occupancy measure. Let $\som$ be the state occupancy measure of policy $\pi$, defined as

 \begin{align*}
     \som(s_f) = E_{a \sim \pi(\cdot | s) s \sim q}[\sr(s_f | s, a)]
 \end{align*}

where $q$ is the distribution of initial states. With a slight abuse of notation, we define $$\som(s, a) = \pi(a | s)\som(s)$$.

We then present an upper bound on the divergence between the data set and the state occupancy measure. 

\begin{align*}
    KL(D(s, a) || \som(s,a))
        &= E_{s \sim D}[KL(D(a|s) || \pi(a|s))] + KL(D(s) || \som(s))\\
     &= E_{s\sim D}[KL(D(a|s) || \pi(a|s))] + KL( \int_{s_0} D(s_f| s_0) q(s_0) || \int_{s_0} \sr(s_f | s_0)q(s_0))]\\
    (\textit{Jensen's Inquality}) &\leq E_{s\sim D}[KL(D(a|s) || \pi(a|s))] + E_{s_0\sim q}[KL(D(s_f| s_0) || \sr(s_f | s_0))]\\
    (\textit{Jensen's Inquality}) &\leq E_{s\sim D}[KL(D(a|s) || \pi(a|s))] + E_{s_0\sim q, a_0 \sim \pi(\cdot | s_0)}[KL(D(s_f| s_0) || \sr(s_f | s_0,a_0))]\\
\end{align*}

In the case that the MDP is ergodic, this can be simplified further. If this is the case, then the occupancy measure for a sufficiently large $\gamma$ approaches the same distribution regardless of the starting state \citep{chen2022learninginfinitehorizonaveragerewardmarkov}. As a result, the above inequality is approximately true for any $q$, because the choice of $q$ does not affect the distribution of $D$ or $\sr$. We choose $q = D$. Then, 

\begin{align}\label{result:1}
    KL(D(s, a) || \som(s,a))
        &\lessapprox E_{s\sim D}[KL(D(a|s) || \pi(a|s))] + E_{s\sim D, a \sim \pi(\cdot | s)}[KL(D(s_f| s) || \sr(s_f | s,a))]
\end{align}

The left term is a standard behavior cloning loss. Recall that TD3 and ReBRAC both learn Gaussian policies with a fixed variance $\sigma^2$ and mean $\mu_\theta$ \citep{TD3, tarasov2023revisitingminimalistapproachoffline}. Then if the action space has dimension $d_A$, we have

\begin{align*}
    KL(D(a|s) || \pi(a|s)) &= E_{a\sim D(\cdot|s)}[-\log(\pi(a|s))] - H(D(a|s)) \\
    &= E_{a\sim D(\cdot|s)}\left[\frac{||a - \mu_\theta(s) ||^2}{2\sigma^2}\right] + d_A \log(\sigma) + \frac{d_A}{2} \log (\pi) - H(D(a|s)) \\
    &\leq E_{a\sim D(\cdot|s)}\left[\frac{||a - \mu_\theta(s) ||^2}{2\sigma^2}\right] + d_A \log(\sigma) + \frac{d_A}{2} \log (\pi) \\
\end{align*}

We can see that this is identical to the TD3 behavior cloning loss, up to constant terms. \footnote{Note that the $ \frac{d_A}{2} \log (\pi)$ term is referring to the mathematical constant $\pi=3.14...$, not the policy $\pi$. This term does not depend on the policy}

The right term of ~\ref{result:1} is effectively a behavior cloning term on the state distribution. The KL divergence for this right term can easily be upper bounded using the standard diffusion model loss. As previously described, we learn a Bellman Diffusion Model to represent $\sr(s_f | s, a)$. We then minimize the imitation loss, backpropagating through $\sr(s_f | s, a)$ to train the policy.

Combining these two results, we find

\begin{align*}
    KL(D(s, a) || \som(s,a))
        &\lessapprox E_{s\sim D}[E_{a\sim D(a|s)}\left[\frac{||a - \mu_\theta(s) ||^2}{2\sigma^2}\right] \\& + (K-1)E_{i, \epsilon, s_f \sim D(s_f | s), a \sim \pi( \cdot |s)}[\eta_i || \epsilon - \epsilon_\theta(s_{f_i}, s, a_\pi, i)||^2]] + C
\end{align*}

where $C$ is a constant that does not depend on the policy. 

This is the regularization loss used by TD3-SBC, up to constants scaling the balance between the action and state regularization losses.

\section{Experiment Details}

All experiments were performed on an internal cluster, using a mixture of  RTX A4500 and GeForce RTX A4500 GPUs. Each training process was assigned 1 GPU, and took between 12 and 24 hours to complete, requiring approximately 900 GPU hours total across all runs. Hyperparameter tuning took a similar amount of compute. Additionally, several previous iterations of the algorithm were tested, but the compute cost of these experiments was not tracked. 

Our implementation was based on OfflineRLKit, and evaluated on the D4RL datasets \citep{offinerlkit, fu2021d4rldatasetsdeepdatadriven}.

For the most part, we reused the hyperparameters from ReBRAC, which were found to work well. We tuned $w_s$, the state regularization weight, $w_a$, the action regularization weight, and the actor learning rate. We found that it was necessary to tune the actor and diffusion learning rates because the gradient from the diffusion model is stochastic, unlike the gradient from the Q function and the behavior cloning loss. Reducing the learning rate made convergence much more stable. 

In order to reduce the number of hyperparameter search runs, we first turned off the value function training, and did a grid search over $w_s$, and the actor and diffusion model learning rates. The idea was to first find the optimal mix of state and action regularization, and the learning rate needed for that mix to be stable. Then, we searched for the optimal weighting of the value function, given a fixed mixture of state and action regularization. Since we searched three values each for lr and state regularization and five values for Q value weight, we reduced the number of runs needed from 3*3*5 = 45, to 3*3 + 5 = 14, a nearly 70\% reduction. 

We refer to ReBRAC's parameters in the following way:

\begin{align*}
	& w_a^0: \text{Action regularization coefficient used by ReBRAC} \\
	& lr_\pi^0: \text{Actor learning rate used by ReBRAC}  \\
	& lr_Q^0: \text{Critic learning rate used by ReBRAC} \\
\end{align*}

The values we searched for in the initial imitation learning runs were parameterized as follows:

\begin{align*}
	w_{s/a}&: \text{Relative state regularization coeffiecient} \\
	\omega &: \text{Actor and diffusion model learning rate reduction factor} \\
	w_a = w_a^0&: \text{Our action regularization coefficient} \\
	w_s = w_{s/a} w_a^0&: \text{Our state regularization coefficient} \\
	lr_\pi = \frac{lr_\pi^0}{\omega}&: \text{Our actor learning rate} \\
	lr_\sr = \frac{lr_Q^0}{\omega}&: \text{Our diffusion model learning rate} 
\end{align*}

Our hyperparameter search used values [1,10,100] for both $w_{s/a}$ and $\text{lr\_reduce}$. For $w_{s/a}$, we found that values < 1 were small enough to have effectively no impact. For $\div$, we only explored values greater than 1 because the use of the diffusion model increased variance, sometimes requiring lower learning rates to maintain stability. Convergence was fast enough that we never needed to increase learning rates. 

For the offline RL portion, we used the following parameterizations:

\begin{align*}
	w_{q} &: \text{Regularization reduction factor}\\
	w_a = \frac{w_a^0}{w_{q}}&: \text{Our action regularization coefficient}\\
	w_s = \frac{w_{s/a} w_a^0}{w_{q}}&: \text{Our state regularization coefficient}
\end{align*}

We searched the values [1,2,4,10,100] for $w_q$. 

We found the following values to be optimal: 

\begin{table}[ht]
\caption{Tuned hyperparameters for TD3-SBC}
\begin{tabular}{lllll}
\toprule
\multicolumn{1}{c}{\textbf{Task Name}} & \multicolumn{1}{c}{$w_{s/a}$} & \multicolumn{1}{c}{$\omega$} & \multicolumn{1}{c}{$w_Q$} & \\
                          \hline
halfcheetah-medium        & 100       & 10            & 1     &  \\
halfcheetah-expert        & 1         & 1             & 1     &  \\
halfcheetah-medium-expert & 1         & 1             & 1     &  \\
halfcheetah-medium-replay & 10        & 10            & 10    &  \\
\hline
hopper-medium             & 100       & 100           & 2     &  \\
hopper-expert             & 1         & 10            & 1     &  \\
hopper-medium-expert      & 100       & 100           & 1     &  \\
hopper-medium-replay      & 100       & 100           & 10    &  \\
\hline
walker2d-medium           & 1         & 100           & 2     &  \\
walker2d-expert           & 10        & 100           & 1     &  \\
walker2d-medium-expert    & 1         & 1             & 1     &  \\
walker2d-medium-replay    & 100       & 100           & 4     & 
\end{tabular}
\end{table}

In terms of the original hyperparameters, these reduce to: 

\begin{table}[ht]
\caption{Final hyperparameters for TD3-SBC}
\begin{tabular}{llllll}

\toprule
\multicolumn{1}{c}{\textbf{Task Name}} & \multicolumn{1}{c}{$w_{a}$} & \multicolumn{1}{c}{$w_s$} & \multicolumn{1}{c}{$lr_\pi$} & \multicolumn{1}{c}{$lr_\text{diffusion}$} \\
                          \toprule
halfcheetah-medium & 0.001 & 0.1 & 0.0001 & 0.0001 \\ 
halfcheetah-expert & 0.01 & 0.01 & 0.001 & 0.001 \\ 
halfcheetah-medium-expert & 0.01 & 0.01 & 0.001 & 0.001 \\ 
halfcheetah-medium-replay & 0.001 & 0.01 & 0.0001 & 0.0001 \\ 
hopper-medium & 0.005 & 0.5 & 1e-05 & 1e-05 \\ 
hopper-expert & 0.1 & 0.1 & 0.0001 & 0.0001 \\ 
hopper-medium-expert & 0.1 & 10.0 & 1e-05 & 1e-05 \\ 
hopper-medium-replay & 0.005 & 0.5 & 1e-05 & 1e-05 \\ 
walker2d-medium & 0.025 & 0.025 & 1e-05 & 1e-05 \\ 
walker2d-expert & 0.01 & 0.1 & 1e-05 & 1e-05 \\ 
walker2d-medium-expert & 0.01 & 0.01 & 0.001 & 0.001 \\ 
walker2d-medium-replay & 0.0125 & 1.25 & 1e-05 & 1e-05 \\
\bottomrule
\end{tabular}
\end{table}

\section{Additional Experimental Results}

In addition to our offline learning experiments, we also tried removing the rewards and Q function, and comparing State Behavior Cloning (SBC) alone against ordinary behavior cloning and SMODICE, an offline imitation learning algorithm \citep{ma2022versatileofflineimitationobservations}. We find that SBC shows strong performance, outperforming BC and SMODICE on five of the twelve environments. Interestingly, SBC appears to be more stable than SMODICE, which collapsed and achieved near-zero reward on 3 of the 4 walker environments, and even crashed on walker2d-medium-expert from numerical overflows.

\begin{table}[ht]
\caption{Imitation learning experimental results on the D4RL dataset. Reported values are the last-iterate return averaged across all training seeds. The $\pm$ symbol represents the standard deviation across seeds. }
\begin{tabular}{llllll}
\toprule
\multicolumn{1}{c}{\textbf{Task Name}} & \multicolumn{1}{c}{BC} & \multicolumn{1}{c}{SMODICE} & \multicolumn{1}{c}{SBC} \\
                          \toprule
halfcheetah-medium & 42.51 $\pm$ 0.38 & \textbf{55.99 $\pm$ 4.19} & 42.21 $\pm$ 0.37 \\ 
halfcheetah-expert & 93.17 $\pm$ 0.11 & \textbf{94.10 $\pm$ 1.05} & 93.05 $\pm$ 0.20 \\ 
halfcheetah-medium-expert & 66.83 $\pm$ 7.08 & \textbf{81.96} \textbf{$\pm$ 5.22} & 69.07 $\pm$ 7.88  \\ 
halfcheetah-medium-replay & 33.97 $\pm$ 4.84 & \textbf{88.45 $\pm$ 2.73} & 36.67 $\pm$ 2.06 \\ 
\midrule
hopper-medium & {53.00 $\pm$ 0.57} & {57.01 $\pm$ 3.88} & {\textbf{65.40 $\pm$ 7.88}}  \\ 
hopper-expert & 109.46 $\pm$ 1.80 & 111.12 $\pm$ 0.20 & \textbf{111.46 $\pm$ 0.51}  \\ 
hopper-medium-expert & 56.20 $\pm$ 12.79 & 61.70 $\pm$ 7.91 & \textbf{86.10 $\pm$ 16.55}  \\ 
hopper-medium-replay & 19.61 $\pm$ 2.63 & \textbf{69.18 $\pm$ 24.71} & 43.93 $\pm$ 25.58 \\ 
\midrule
walker2d-medium & 71.72 $\pm$ 4.46 & 0.30 $\pm$ 0.53 & \textbf{73.73 $\pm$ 3.61}  \\ 
walker2d-expert & \textbf{108.51 $\pm$ 0.43} & 107.62 $\pm$ 0.34 & \textbf{108.51 $\pm$ 0.06}  \\ 
walker2d-medium-expert & 88.38 $\pm$ 14.41 & crashed & \textbf{95.55 $\pm$ 13.15}  \\ 
walker2d-medium-replay & \textbf{33.20 $\pm$ 12.52} & 4.46 $\pm$ 4.97 & 33.11 $\pm$ 14.71  \\
\bottomrule
\end{tabular}
\end{table}
For these experiments, we used the same hyperparameters as the offline learning experiments, but omitted the $Q$ function from the policy loss. For SMODICE, we used the author's original implementation with its original hyperparameters.

\end{document}